\documentclass{article}

\usepackage{microtype}
\usepackage{graphicx}
\usepackage{subfigure}
\usepackage{booktabs} % for professional tables

\usepackage{hyperref}

\usepackage[accepted]{icml2023}

\usepackage{amsmath}
\usepackage{amssymb}
\usepackage{mathtools}
\usepackage{amsthm}

\usepackage{amsmath}
\usepackage{amssymb}
\usepackage{mathtools}
\usepackage{amsthm}

\usepackage{url}
\usepackage{graphics}
\usepackage{microtype}
\usepackage{graphicx}
\usepackage{subfigure}
\usepackage{booktabs} % for professional tables
\usepackage{xcolor}
\usepackage{amsmath}
\usepackage{amssymb}
\usepackage{amsthm}
\usepackage{bm}
\usepackage{graphicx}
\usepackage{wrapfig}
\usepackage{diagbox}
\usepackage{multirow}
\usepackage{enumitem}
\usepackage{MnSymbol}
\usepackage{dsfont}
\usepackage{boldline, arydshln}
\usepackage{cuted}
\usepackage{hhline}
\usepackage{caption}
\usepackage{lipsum}
\usepackage{pifont}
\usepackage{makecell}
\usepackage{caption}

\usepackage[capitalize,noabbrev]{cleveref}

\theoremstyle{plain}

\theoremstyle{definition}

\theoremstyle{remark}

\def\onedot{.}
 
\def\ie{\emph{i.e}\onedot}

\usepackage{graphicx}
\newcommand{\cmark}{\ding{51}}
\newcommand{\xmark}{\ding{55}}

\usepackage[textsize=tiny]{todonotes}

\icmltitlerunning{MonoNeRF: Learning Generalizable NeRFs from Monocular Videos without Camera Pose}

\begin{document}

\twocolumn[
% \icmltitle{Multiplane NeRF-Supervised 
% Disentanglement of \\ Depth and Camera Pose from Videos}
\icmltitle{MonoNeRF: Learning Generalizable NeRFs from Monocular Videos \\ without Camera Poses}



\icmlsetsymbol{equal}{*}

\begin{icmlauthorlist}
\icmlauthor{Yang Fu}{yyy}
\icmlauthor{Ishan Misra}{comp}
\icmlauthor{Xiaolong Wang}{yyy}
\end{icmlauthorlist}

\icmlaffiliation{yyy}{University of California, San Diego}
\icmlaffiliation{comp}{FAIR, Meta AI}

\icmlcorrespondingauthor{Xiaolong Wang}{xiw012@ucsd.edu}


\icmlkeywords{Machine Learning, ICML}

\vskip 0.15in

\renewcommand\twocolumn[1][]{#1}%
% \maketitle
% \vspace{-10mm}%
\begin{center}
    \centering
	\includegraphics[width=0.9\textwidth]{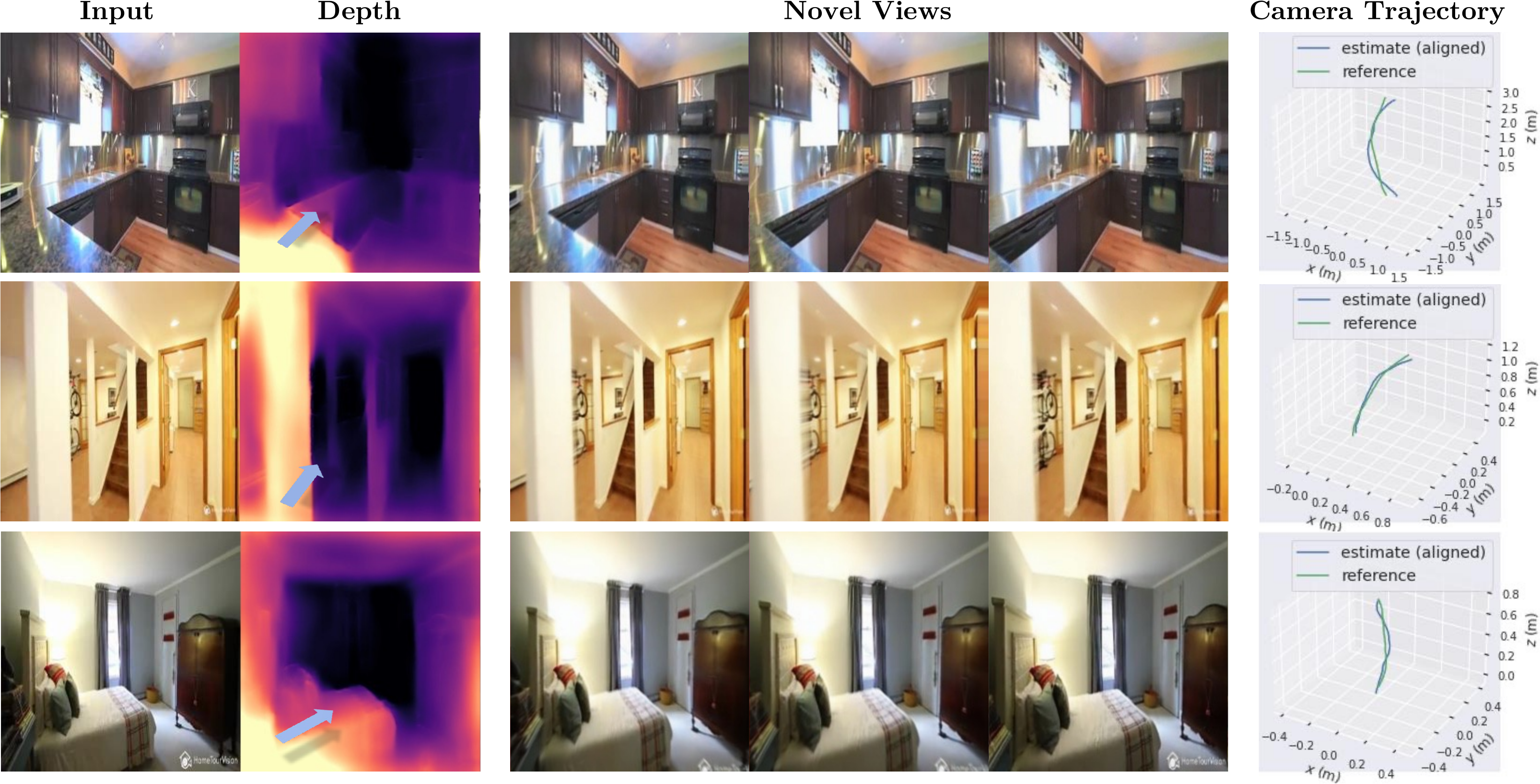}
	\vspace{-2mm}
	\captionof{figure}{We learn a MonoNeRF from monocular videos that can be applied to \textbf{depth estimation, novel view synthesis}, and \textbf{camera pose estimation}. 
 % More video examples can be found at supplementary materials.
 }
    % Our method first learns a generalizable scene representation from different scenes. The learnt generalizable scene representation can be utilized to (a) extract scene priors without per-scene optimization. Given the scene prior, we perform the per-scene optimization in a more accurate and efficient way leading to (b) high-quality surface reconstruction and (c) more realistic textured meshes.}
    \label{fig:teaser}
    \vspace{2mm}%
\end{center}%

% \begin{figure}[t]
%     \centering
%     \vspace{-2mm}
% 	\includegraphics[width=0.48\textwidth]{figs/teaser_small.pdf}
% 	\vspace{-7mm}
% 	\caption{MonoNeRF learns a generalizable NeRF representation from monocular videos that can be applied to depth estimation, novel view synthesis, and camera pose estimation.}
%  % More video examples can be found at supplementary materials.
%     % Our method first learns a generalizable scene representation from different scenes. The learnt generalizable scene representation can be utilized to (a) extract scene priors without per-scene optimization. Given the scene prior, we perform the per-scene optimization in a more accurate and efficient way leading to (b) high-quality surface reconstruction and (c) more realistic textured meshes.}
%     \label{fig:teaser}
%     % \vspace{-7mm}%
% \end{figure}

]


\printAffiliationsAndNotice{} % otherwise use the standard text.

\begin{abstract}
We propose a generalizable neural radiance fields - \textit{MonoNeRF}, that can be trained on large-scale monocular videos of moving in static scenes without any ground-truth annotations of depth and camera poses. \textit{MonoNeRF} follows an Autoencoder-based architecture, where the encoder estimates the monocular depth and the camera pose, and the decoder constructs a Multiplane NeRF representation based on the depth encoder feature, and renders the input frames with the estimated camera. The learning is supervised by the reconstruction error. Once the model is learned, it can be applied to multiple applications including depth estimation, camera pose estimation, and single-image novel view synthesis. More qualitative results are available at: \url{https://oasisyang.github.io/mononerf}.

% We propose to perform self-supervised disentanglement of depth and camera pose from large-scale videos. We introduce an Autoencoder-based method to reconstruct the input video frames for training, without using any ground-truth annotations of depth and camera. The model encoders estimate the monocular depth and the camera pose. The decoder then constructs a Multiplane NeRF representation based on the depth encoder feature, and renders the input frames with the estimated camera. The learning is supervised by the reconstruction error, based on the assumption that the scene structure does not change in short periods of time in videos. Once the model is learned, it can be applied to multiple applications including depth estimation, camera pose estimation, and single image novel view synthesis. We show substantial improvements over previous self-supervised approaches on all tasks and even better results than counterparts trained with camera ground-truths in some applications. More results can be found in: \url{https://icml-submission.github.io/}. Our code will be made publicly available. 
\end{abstract}

\section{Introduction}
\label{sec:intro}

The Neural Radiance Fields (NeRF) have been successfully applied in many applications on not just view synthesis~\cite{mildenhall2020nerf,martin2021nerf}, but also scene and object reconstruction~\cite{wang2021neus,yariv2021volume,zhang2021ners}, semantic understanding~\cite{Zhi:etal:ICCV2021} and robotics~\cite{li20223d,simeonov2022se}. While these results are encouraging, constructing NeRF requires accurate ground-truth camera poses and the learned NeRF is specific to only one scene in most cases. This usually takes a significant amount of time for training and it also limits the applications in large-scale unconstrained videos. 

To accelerate the optimization process of NeRF, more recent focus has been spent on learning generalizable NeRF~\cite{yu2021pixelnerf,li2021mine,chen2021mvsnerf} which is first trained on a dataset with multiple scenes, and then fine-tuned on each individual scene. The learning of a generalizable representation provides a prior that not only accelerates the optimization (i.e., fine-tuning) process, but also allows reconstruction and view synthesis with only a few view inputs instead of using dense views. However, all these approaches still require training on datasets with given camera poses. While there are also studies on training NeRF without camera poses~\cite{wang2021nerf,wang2021nerfmm}, all these efforts are focused on training a NeRF on a single scene instead of generalizing across scenes. One fundamental reason behind this is: it is very challenging to perform calibrations across scenes in a self-supervised way.

% Multiple continuous frames from the video can reconstruct multiplane images for a given view~\cite{tucker2020single}. 

In this paper, we propose a novel generalization NeRF called \textit{MonoNeRF}, which can be learnt from monocular videos of moving in static scenes without using any camera ground truths. Our key insight is that,  real-world videos often come with slow camera changes (continuity) instead of presenting diverse viewpoints. With this observation, we propose to train an Autoencoder-based model on large-scale real-world videos. Given the input video frames, our framework uses a depth encoder to perform monocular depth estimation for each frame (which is encouraged to be consistent), and a camera pose encoder to estimate the relative camera pose between every two consecutive frames. The \textbf{depth encoder feature} and the \textbf{camera pose} are the intermediate disentangled representations. For each input frame, we construct a NeRF representation with the depth encoder feature and render it to decode another input frame based on the estimated camera pose. We train the model with the reconstruction loss between the rendered frames and the input frames. However, using a  reconstruction loss alone can easily lead to a trivial solution as the estimated monocular depth, camera pose, and the NeRF representation are not necessarily on the same scale. One \textbf{key technical contribution} we propose is a novel scale calibration method during training to align these three representations. The advantages of our framework are: (i) Unlike NeRF, it does not need 3D camera pose annotations (e.g., computed via SfM); (ii) It generalizes training on a large-scale video dataset, which leads to better transfer.

At test time, the learned representations can be applied to multiple downstream tasks including: (i) monocular depth estimation from a single RGB image; (ii) camera pose estimation; (iii) single-image novel view synthesis. \emph{We conduct all experiments on indoor scenes in this paper,} as shown in Figure~\ref{fig:teaser}.  For depth estimation, we train on Scannet~\cite{dai2017scannet}. Our method significantly improves over previous self-supervised depth estimation approaches not only on the Scannet test set and also generalizes to NYU Depth V2~\cite{Silberman_ECCV12} better. For camera pose estimation, we use RealEstate10K~\cite{zhou2018stereo} following~\cite{lai2021video} and consistently achieve much better performance compared to previous approaches. For novel view synthesis from a single image input, we estimate the monocular depth using the depth encoder, construct the multiplane NeRF, and then render another view with a given camera. On RealEstate10K~\cite{zhou2018stereo}, our approach significantly improves over methods that learn without camera ground-truth and also outperform recent methods that learn with the ground-truth cameras~\cite{wiles2020synsin}. To our knowledge, our method is \textbf{the first work that learns neural radiance fields on a large-scale dataset without camera ground truth}.

\section{Related Work}
\label{sec:related}

\textbf{Novel View Synthesis and Neural Radiance Fields.} Learning-based novel view synthesis has been a long stand task. Researchers have studied on using explicit 3D representations including voxels~\cite{jimenez2016unsupervised,kar2017learning,tulsiani2017multi,sitzmann2019deepvoxels,tung2019learning,nguyen2019hologan}, depth maps~\cite{wiles2020synsin,rockwell2021pixelsynth} and multiplane image~\cite{zhou2018stereo,srinivasan2019pushing,tucker2020single,li2021mine} for view synthesis. For example, \citet{wiles2020synsin} proposed to infer the depth map from the input image as an intermediate representation and perform rendering from another view for synthesis. Instead of using a single depth map, multiplane image (MPI) representation is utilized to explicitly model the occluded contents during view synthesis~\cite{tucker2020single}. Besides explicit 3D representations, recent work on using implicit representations have shown superior performance in view synthesis~\cite{sitzmann2019scene,niemeyer2020differentiable}.  Following this line of research, NeRF and its subsequent works~\cite{yu2021pixelnerf,trevithick2021grf,martin2021nerf,schwarz2020graf,wang2021nerf,meng2021gnerf} have even achieved photo-realistic rendering results. 
While the original formulation is restricted to one single instance with the provided camera, recent extensions have made it available to generalize to multiple instances with camera ground-truths~\cite{yu2021pixelnerf,trevithick2021grf,li2021mine,chen2021mvsnerf,wang2021ibrnet,wang2022attention,venkat2023geometry} or training on \textbf{a single scene} without camera ground-truths~\cite{wang2021nerf,meng2021gnerf}. For example, \citet{yu2021pixelnerf} propose to leverage an encoder network to model the scene priors and train a generalizable NeRF on diverse scenes with given camera poses. On the other hand, \citet{wang2021nerf} shows that the camera pose can be jointly optimized as learnable parameters with NeRF training. However, this approach only works on training NeRF for a single scene. None of the previous works can generalize to training on large-scale data and without cameras at the same time. 

\textbf{Disentangled representations.} Disentangled representations aim to decompose complex visual data into several lower-dimensional individual factors that control different types of attributes. Common approaches to achieve disentanglement include using Generative Adversarial Networks~\cite{chen2016infogan, huang2018multimodal, karras2019style, lee2020drit++, zhu2018visual} and Autoencoders~\cite{jha2018disentangling, liu2020learning, park2020swapping, pidhorskyi2020adversarial}. For instance, \citet{park2020swapping} proposed an Autoencoder to disentangle texture from the structure by enforcing one component to encode co-occurrent patch statistics across different parts of the image. Besides learning from images, recently researchers have looked into using the temporal continuity in videos for learning disentangled representations~\cite{denton2017unsupervised,minderer2019unsupervised,wiles2018self,xue2016visual,lai2021video}. The most related work to our method is VideoAE ~\citep{lai2021video}, where an autoencoder network is proposed to disentangle the static 3D scene structure and camera motion from videos. However, their 3D structure is represented by deep voxel features, which cannot reveal the explicit scene geometric structure. The proposed MonoNeRF is able to directly infer depth as the scene representation, which can be directly used as a downstream application.

\textbf{Self-Supervised Depth Estimation.} Single image depth estimation has been widely studied in a supervised learning setting~\cite{eigen2014depth,laina2016deeper,kendall2017end}. However, with the absence of ground-truth depth or camera pose in most real-world data, self-supervised approaches using image reconstruction as the training signal without relying on neither depth nor camera annotations are proposed~\cite{zhou2017unsupervised,vijayanarasimhan2017sfm,yin2018geonet,yang2018lego,mahjourian2018unsupervised,gordon2019depth,li2020unsupervised}. In this paper, we also follow the setting on learning without both depth and camera ground-truths and apply it on indoor scenes. Different from previous approaches, we show that depth can be learned by rendering with multiplane NeRF, which not only significantly improves depth estimation, but also allows better camera estimation and novel view synthesis results. Our work is also related to self-supervised learning of visual representations from videos~\cite{agrawal2015learning,han2019video,misra2016shuffle,wang2015unsupervised,wang2019learning,jabri2020space}. However, instead of focusing on learning representations for recognition tasks, our work is more focused on scene geometric understanding for tasks including camera pose estimation, depth estimation, and novel view synthesis.

\section{Method}
\label{sec:method}

\begin{figure*}[t]
    \centering
	\includegraphics[width=0.95\textwidth]{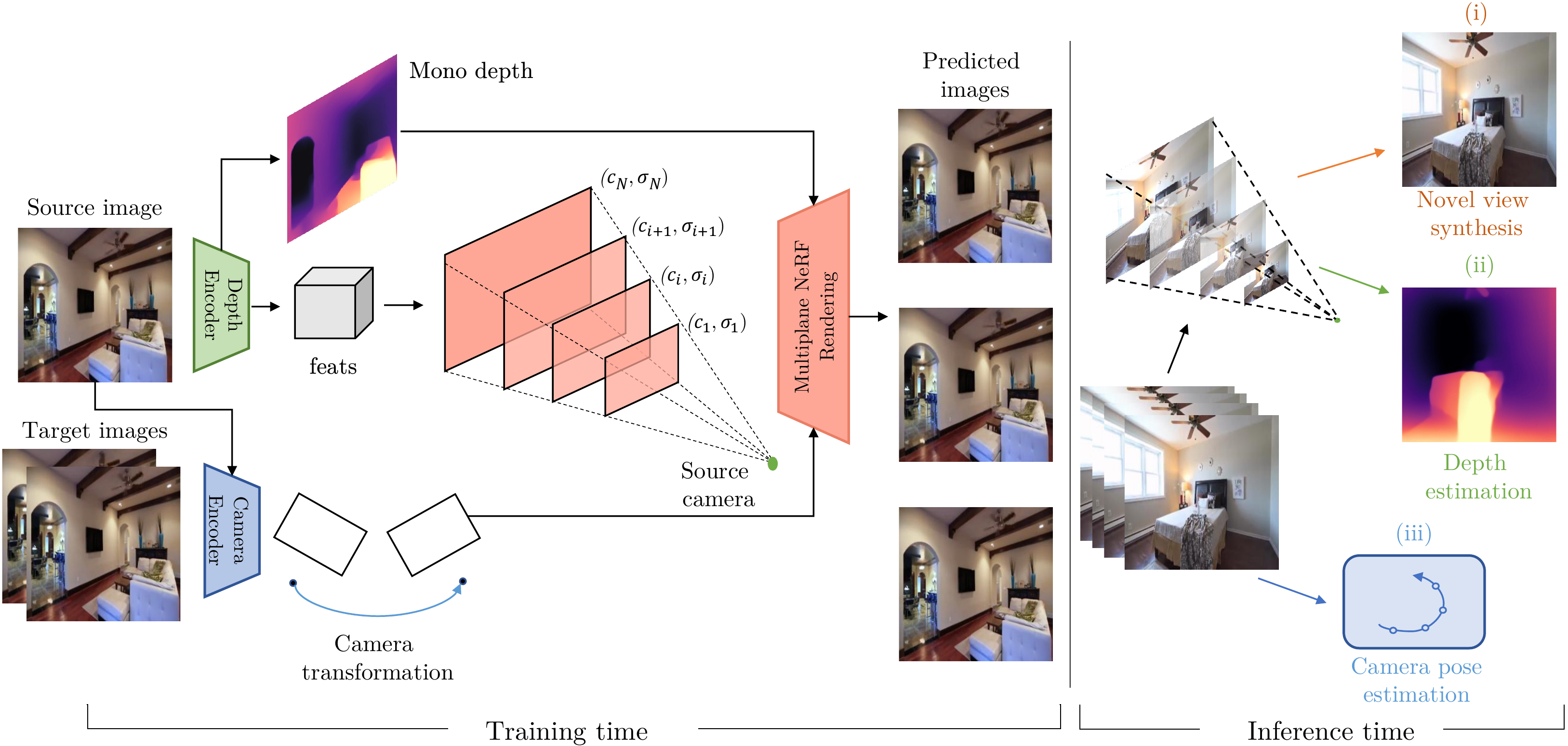}
	\vspace{-2mm}
	\caption{\textbf{Overview of proposed MonoNeRF}. Given a short clip of video, the camera encoder and depth encoder disentangle it into depth maps, neural representations, and relative camera trajectory. The Multiplane NeRF is utilized as the decoder to generate the target images according to the estimated camera pose. During training, the model is supervised via the reconstruction loss between the input frames and the generated ones. During testing, three downstream tasks, \ie camera pose estimation, depth estimation, and novel view synthesis can be achieved within a single model.}
	\label{fig:approach}
	\vspace{-2mm}
\end{figure*}

% \vspace{-4mm}
% In this work, we aim to learn disentangled 3D representations from videos in a self-supervised manner (\textbf{no camera pose and depth ground-truths}) in an autoencoder fashion. The inputs to our model are video frames (3 frames in our experiments) that are nearby in a short period of time. The video frames are processed with the depth encoder and the camera pose encoder for the depth estimation and camera trajectory estimation respectively. In the decoding process, we construct the multiplane NeRF representation using the depth encoder feature and render using the estimated cameras. We minimize the reconstruction loss between the rendered frames and the input frames to learn the full model. Our model learns disentanglement of the intermediate representations including the \textbf{depth feature} (which is used to predict depth) and the \textbf{camera pose}. We introduce the encoding process in sections~\ref{3.1} and~\ref{3.2}, the decoding process in section~\ref{3.3}, and the training details in section~\ref{3.4}.

In this work, MonoNeRF aims to learn generalizable NeRF representations from monocular videos in a self-supervised manner (\textbf{no camera pose and depth ground-truths}) in an autoencoder fashion. As multiple continuous frames from the video can reconstruct multiplane images for a given view~\cite{tucker2020single}, we adopt a variant of NeRF~\cite{li2021mine} which combines the discrete multiplane images into NeRF to create continuous multiplane neural radiance fields in our approach. The inputs to our model are video frames (3 frames in our experiments) that are nearby in a short period of time. The video frames are processed with the depth encoder and the camera pose encoder for the depth estimation and camera trajectory estimation respectively. In the decoding process, we construct the multiplane NeRF representation using the depth encoder feature and render using the estimated cameras. We minimize the reconstruction loss between the rendered frames and the input frames to learn the full model. MonoNeRF learns the disentanglement of the intermediate representations including the \textbf{depth feature} (which is used to predict depth) and the \textbf{camera pose}. We introduce the encoding process in sections~\ref{3.1} and~\ref{3.2}, the decoding process in section~\ref{3.3}, and the training details in section~\ref{3.4}.

\vspace{-0.5em}
\subsection{Camera Pose Encoder}\label{3.1}
\vspace{-0.5em}
The camera pose encoder predicts the relative camera transformation between two input frames as shown at the bottom of  Fig~\ref{fig:approach} (blue box). Specifically, given a source frame $I_s$ and a target frame $I_t$ as inputs, it computes the rotation matrix and translation matrix w.r.t the source view image. For an input sequence during training, we use the middle frame as the source view image and take the remaining frames before and after as target images. We follow the ResNet~\cite{he2016deep} architecture to design our encoder, which takes both frames as inputs (\ie, stacked along the channel dimension leading to six input channels) and outputs a $6$-dim vector as the 3D rotation and translation parameters. We formulate the camera encoder as,
 \begin{equation}
% \begin{aligned}
\mathbf{T}_{s \rightarrow t} := [R, \boldsymbol{t}]= \mathcal{F}_{\mathrm{traj}}([I_s, I_t])
\label{equ:cam}
\end{equation}
The estimated camera poses for all target images can construct a trajectory and then be used for target view synthesis in the decoder which will be discussed later.

\subsection{Monocular Depth Encoder}\label{3.2}

We design a separate encoder for monocular depth estimation from each single input frame, as shown in the upper part of Fig.~\ref{fig:approach} (green box). We adopt the network architecture from MnasNet~\cite{tan2019mnasnet} as the depth encoder network, which extracts feature maps with different resolution scales to predict the depth map. We formulate the depth encoder as,
\begin{equation}
% \begin{aligned}
\mathbf{D}_{s} = \mathcal{F}_{\mathrm{dep}}(I_s)
\label{equ:depth}
\end{equation}
Note that the raw output $\mathbf{D}_{s}$ is the disparity map and needs to be converted to the depth map. The output monocular depth map is used as the intermediate representation to guide the construction of Multiplane NeRF.

\subsection{Multiplane NeRF based Decoder}\label{3.3}

The disentanglement is learned via back-propagation from the differentiable decoder. To enable the optimization, we assume that the input video frames are taken from a short range of time scales, and the scene structure remains the same. This assumption provides supervision for our method to construct a Multiplane NeRF representation from a single image in our decoder, and use this representation to render the outputs. We first introduce the multiplane image representation, and then illustrate how to combine it with NeRF to perform rendering in our framework.

% \vspace{-1.5mm}
\paragraph{Multiplane Images.} 
We first review Multiplane Images (MPIs)~\cite{zhou2018stereo}, where an image is represented by a set of parallel planes of RGB-$\alpha$, $\{(c_{i}, \alpha_{i})\}_{i=1}^D$, where $c_{i} \in\mathbb{R}^{H \times W \times 3}$ are RGB values, $\alpha_{i} \in\mathbb{R}^{H \times W \times 1}$ are the alpha values and $D$ is the number of planes. Each plane corresponds to a specific disparity (inverse of depth) value $d_{i}$ uniformly sampled from a predefined range$[d_\mathrm{min}, d_\mathrm{max}]$. Given the rotation matrix $R$ and translation matrix $\boldsymbol{t}$ from target to source view and the intrinsics matrix for source and target views $K_{s},  K_{t}$, we can generate the target-view image $\hat{\mathbf{I}}_{t}$ and the disparity map $\hat{\mathbf{D}}_{s}$ via the following steps. We use $\mathbf{D}$ to denote the monocular depth directly estimated from the network, and $\hat{\mathbf{D}}$ to denote the depth generated by rendering from MPI. First, the warping operation for the $i$-th plane from target to source view can be formulated as the following,
\begin{equation}
% \begin{aligned}
\begin{bmatrix}u_s \\ v_s \\ 1\end{bmatrix}
\sim K_{s} \bigl(R - \boldsymbol{t} \boldsymbol{n}^T d_i \bigr) (K_{t})^{-1} \begin{bmatrix}u_{t} \\v_{t} \\1\end{bmatrix}
\label{equ:warp}
\end{equation}
where $\boldsymbol{n}$ is the norm vector of the $i$-th plane and $[u_s, v_s]$, $[u_t, v_t]$ are coordinates in the source and target views respectively. The MPI representation of the target view can be obtained by warping each layer from the source viewpoint to the desired target viewpoint using Eq.~\ref{equ:warp}. 
Then, the MPI representation under target view $(c_{i}^\prime, \alpha_{i}^\prime)$ can be described as,
\begin{equation}
% \begin{aligned}
c_{i}^\prime(u_t, v_t)  = c_{i}(u_s, v_s) \qquad
\alpha_{i}^\prime(u_t, v_t)  = \alpha_{i}(u_s, v_s)
\label{equ:sam}
\end{equation}
Finally, the RGB image and the disparity map under both the source view and target view can be obtained via the same compositing procedure proposed in~\cite{zhou2018stereo}, 

\begin{equation}
\small
\left\{
\begin{aligned}
&\hat{\mathbf{I}}_{s} \ =& \sum^{D}_{i=1} \bigl( c_i \alpha_i^\prime \prod_{j=i+1}^{D} (1 - \alpha_j) \bigr) \\
&\hat{\mathbf{D}}_{s} =& \sum^{D}_{i=1} \bigl( d_{i} \alpha_i \prod_{j=i+1}^{D} (1 - \alpha_j) \bigr) 
\end{aligned}
\right.
\label{equ:mpi_render}
\end{equation}

\begin{equation}
\left\{
\begin{aligned}
&\hat{\mathbf{I}}_{t} \ =& \sum^{D}_{i=1} \bigl( c_i^\prime \alpha_i^\prime \prod_{j=i+1}^{D} (1 - \alpha_j^\prime) \bigr) \\
&\hat{\mathbf{D}}_{t} =& \sum^{D}_{i=1} \bigl( d_{i} \alpha_i^\prime \prod_{j=i+1}^{D} (1 - \alpha_j^\prime) \bigr)
\end{aligned}
\right.
\label{equ:mpi_render_tgt}
\end{equation}

\paragraph{Multiplane NeRF}. Going beyond RGB images, we generalize the representations by introducing NeRF as \cite{li2021mine}, namely Multiplane NeRF. Different from MPI which consists of multiple planes
of RGB-$\alpha$ images at sparse and discrete depths, the Multiplane NeRF achieves continuous representation of 3D scenes by predicting RGB-$\alpha$ images at any arbitrary depth.  Formally, the image is represented by $\{(c_i, \sigma_i)\}_{i=1}^D$, where $\sigma_i$ is the volume density of the $i$-th plane. We follow a similar setting to construct the Multiplane NeRF representation as our decoder to generate the novel view images. 
Specifically, we extract the intermediate representation from the \emph{monocular depth encoder} (gray cube in Fig.~\ref{fig:approach}) as the image feature for $\mathbf{I}_s$. We combine this feature with a disparity level $d_i$ as the inputs for an internal encoder-decoder module, which outputs the RGB image $c_i$ and the density map $\sigma_i$ as a 4-channel map $\{(c_i, \sigma_i)\}$ (multiple orange planes in Fig.~\ref{fig:approach}). We have different planes of $\{(c_i, \sigma_i)\}$ given different disparity $d_i$, and we use  positional encoding to encode each $d_i$. The $i$-th plane for the Multiplane NeRF representation is formulated as,
\begin{equation}
% \begin{aligned}
\{c_{i}, \sigma_i\} = \mathcal{F}_{\mathrm{mpi}}\bigl(\mathbf{I}_s, \mathrm{PE}(d_i)\bigr)
\label{equ:mpi}
\end{equation}
Note we only need to run the depth encoder once to extract the image feature for $\mathbf{I}_s$.  To reconstruct one target view, given the camera trajectory obtained from the \emph{camera pose encoder} (blue module in Fig~\ref{fig:approach}), we first compute the new RGB and density values on the target view $(c_i^\prime, \sigma_{i}^\prime)$ using homography warpping described in Eq.~\ref{equ:warp}, then replace the alpha map $\alpha$ and the compositing operation in Eq.~\ref{equ:mpi_render} by the volume density $\sigma$ and the naive rendering procedure used in~\cite{mildenhall2020nerf} to obtain the image and the disparity map. The advantages of multiplane NeRF over the vanilla NeRF include: (i) it builds the frustum from a single image; (ii) it has a better generalization ability allowing training on large-scale data, which makes it more feasible than NeRF as the decoder in our autoencoder-like architecture.

% \subsubsection{Camera Trajectory Estimation}
% \vspace{-1mm}
\subsection{Supervision with RGB}\label{3.4}
% \vspace{-1.5mm}

Our model is trained in a self-supervised manner by reconstructing multiple video frames as shown in Fig.~\ref{fig:approach}. During training, we select the center frame of $N$-frame clip ($N=3$ in our experiments) as the source view image $\mathbf{I}_s$. We use the depth encoder to estimate the monocular depth $\mathbf{D}_s$ for the source view. We use the camera encoder taking the source view image  $\mathbf{I}_s$ and the target view image $\mathbf{I}_t$ as the inputs to obtain the relative camera pose $(R, \boldsymbol{t})$. Together with the depth encoder feature (gray box in Fig.~\ref{fig:approach}) and the estimated camera,  we can construct the Multiplane NeRF representation and render the target view $(\hat{\mathbf{I}_t}, \hat{\mathbf{D}}_t)$ as the outputs. The autoencoder is supervised by comparing the rendered target image $\hat{\mathbf{I}}_t$ and the ground-truth target image $\mathbf{I}_t$. However, a direct reconstruction objective can easily lead to trivial solutions given both depth and camera ground truths are not provided in training. We propose \textbf{two key technical contributions} including auto-scale calibration and new loss functions to enable successful disentanglement of depth and camera pose. 

\subsubsection{Auto Scale Calibration}
\label{sec:scalecalibration}
Recall that our Multiplane NeRF is built upon a single image, this can lead to the scale ambiguity issue. As explained in~\cite{tucker2020single, li2021mine}, each training sequence can be considered equally valid when we scale down or up the world coordinate by any constant value. To tackle this issue, \cite{li2021mine} and \cite{tucker2020single} propose to use Structure-from-Motion (SfM) to compute \textbf{camera pose} and the \textbf{depth} (sparse point cloud), where both are at the same scale. The calibration procedure is to adjust the camera pose by comparing the depth from SfM and the rendered depth map from Multiplane NeRF. However, the requirement of running SfM in training and testing is time-consuming and it does not always succeed. 

In this paper, we propose to overcome the limits of SfM, and use the encoders to estimate the camera pose $\mathbf{T}_{s \rightarrow t}$ (Eq.~\ref{equ:cam}) and disparity map $\mathbf{D}_s$ (Eq.~\ref{equ:depth}). In this case, none of the \textbf{camera pose} $\mathbf{T}_{s \rightarrow t}$, the \textbf{disparity map} $\mathbf{D}_s$ and the \textbf{NeRF rendered disparity map} $\hat{\mathbf{D}}_s$ are at the same scale initially. We need to calibrate all three together at the same time in the following two steps. 

(i) First, we encourage the rendered disparity map $\hat{\mathbf{D}}_s$ to be consistent with the disparity prediction $\mathbf{D}_s$ by minimizing the $\mathbf{L}1$ distance between them. In detail, we first convert the disparity map into the depth map and then compute the pixel-wise L1 distance between them,
\begin{equation}
% \begin{aligned}
\mathcal{L}_{\mathrm{consist}} = \frac{1}{HW}\sum | \frac{1}{\mathbf{D}_s} - \frac{1}{\hat{\mathbf{D}}_s} |_{1}
\label{equ:depth_loss}
\end{equation}
The above step aligns the rendered depth result with the monocular depth estimation result. 

(ii) Meanwhile, we need to guarantee that the monocular depth estimation and the estimated camera pose are on the same scale. We achieve this goal via applying a photometric reprojection loss~\cite{godard2019digging} between the original source image $\mathbf{I}_s$ and the synthesized source image $\mathbf{I}_{t \rightarrow s}$, obtained by projecting pixels from $\mathbf{I}_t$ onto $\mathbf{I}_s$ given the predicted monocular depth $\mathbf{D}_s$, camera transformation $\mathbf{T}_{s \rightarrow t}$ and the camera intrinsic $\mathbf{K}_s$.
% Besides, to ensure the rendered image $\hat{\mathbf{I}}_{t}$ no longer varies with the scale of the input image, we have to make sure the monocular depth always aligns well with the camera pose. 
% applying a photometric reprojection loss~\cite{godard2019digging} between the original source image $\mathbf{I}_s$ and the synthesized source image $\mathbf{I}_{t \rightarrow s}$, obtained by projecting pixels from $\mathbf{I}_t$ onto $\mathbf{I}_s$ given the predicted monocular depth $\mathbf{D}_s$, camera transformation $\mathbf{T}_{s \rightarrow t}$ and the camera intrinsic $\mathbf{K}_s$.
\begin{equation}
\begin{aligned}
\mathcal{L}_{\mathrm{reproj}} = & \frac{1}{HW}\sum |\mathbf{I}_s - \mathbf{I}_{t \rightarrow s}|_{1}
\end{aligned}
% \vspace{-0.8em}
\label{equ:reproj}
\end{equation}
where $\mathbf{I}_{t \rightarrow s} = \mathbf{I}_{t}\langle \mathrm{proj}(\mathbf{D}_s, \mathbf{T}_{s \rightarrow t}, \mathbf{K}_s)\rangle$.

These two steps can achieve the calibration among the camera pose $\mathbf{T}_{s \rightarrow t}$, the disparity map $\mathbf{D}_s$ and the NeRF rendered disparity map $\hat{\mathbf{D}}_s$ by enforcing the alignment between the synthesized disparity map $\hat{\mathbf{D}}_s$ and the estimated disparity map $\mathbf{D}_s$ as well as the alignment between camera pose $\mathbf{T}_{s \rightarrow t}$ and the estimated disparity map $\mathbf{D}_s$ simultaneously.  

\subsubsection{Loss Functions}
In addition to the calibration, we also adopt three loss functions: RGB L1 loss $\mathcal{L}_{\mathrm{L1}}$, RGB SSIM loss $\mathcal{L}_{\mathrm{ssim}}$ and edge-aware disparity map smoothness loss $\mathcal{L}_{\mathrm{edge}}$ as described in~\cite{tucker2020single}. The RGB L1 loss and SSIM loss~\cite{wang2004image} are defined as,
\begin{equation}
\begin{aligned}
    \mathcal{L}_{\mathrm{L1}} = \frac{1}{HW} \sum |\hat{\mathbf{I}}_t - \mathbf{I}_t| \\
    % \mathcal{L}_{\mathrm{SSIM}} = 1 - \mathrm{SSIM}(\hat{\mathbf{I}}_t, \mathbf{I}_t)
\end{aligned}
\end{equation}
\begin{equation}
\begin{aligned}
    % \mathcal{L}_{\mathrm{L1}} = \frac{1}{HW} \sum |\hat{\mathbf{I}}_t - \mathbf{I}_t| \\
    \mathcal{L}_{\mathrm{SSIM}} = 1 - \mathrm{SSIM}(\hat{\mathbf{I}}_t, \mathbf{I}_t)
\end{aligned}
\end{equation}
Both losses aim at matching the synthesized target image with the ground-truth one. Both $\hat{\mathbf{I}}_t$ and $\mathbf{I}_t$ are RGB images with the size of $H \times W$. Meanwhile, we impose an edge-aware smoothness loss on the synthesized disparity map to align the edge and smoothness region between the disparity map and the original image~\cite{godard2017unsupervised, godard2019digging, tucker2020single, li2021mine}, 
\begin{equation}
    \mathcal{L}_{\mathrm{smooth}} = |\mathbf{\partial}_x \frac{\hat{\mathbf{D}}_s}{\bar{\mathbf{D}}_s}| \exp^{-|\partial_x \mathbf{I}|}  +  |\mathbf{\partial}_y \frac{\hat{\mathbf{D}}_s}{\bar{\mathbf{D}}_s}| \exp^{-|\partial_y \mathbf{I}|}
\end{equation}
where ${\partial}_x$ and ${\partial}_y$ are image gradients and $\bar{\mathbf{D}}_s$ is the mean value of the disparity map $\mathbf{D}_s$. Overall, together with the scale calibration losses, the total is:
\begin{equation}
\begin{aligned}
    \mathcal{L} = \lambda_{\mathrm{L1}} \mathcal{L}_{\mathrm{L1}} + \lambda_{\mathrm{SSIM}} \mathcal{L}_{\mathrm{SSIM}} + \lambda_{\mathrm{smooth}} \mathcal{L}_{\mathrm{smooth}} + \\ \lambda_{\mathrm{consist}} \mathcal{L}_{\mathrm{consist}} + \lambda_{\mathrm{reproj}} \mathcal{L}_{\mathrm{reproj}}
\end{aligned}
\label{equ:loss}
\vspace{-0.5em}
\end{equation}

% \subsubsection{Inference}

\section{Experiments}
\label{sec:exp}

% \vspace{-2mm}

We empirically evaluate MonoNeRF and compare it to the existing approaches on three different tasks: monocular depth estimation, camera pose estimation, and single image novel view synthesis. We perform evaluations on indoor scenes. Compared to outdoor street views, indoor scenes have more structural variance and are more commonly used for evaluating all three tasks together.

% \vspace{-2mm}
\subsection{Implementation Details}\label{4.1}
% \vspace{-2mm}
In the pre-processing step, we resize all images to the resolution of $256 \times 256$ for both training and testing. During training, we randomly sample $3$ frames per sequence with the interval of $5$ as the input to ensure the camera motion is large enough. The number of planes $D$ is set to $64$ and the range of camera frustum is predefined as $[0.2, 20]$. We train our model end-to-end using a batch size of $4$ with an Adam optimizer for $10$ epochs. The initial learning rate is set to $0.0001$ and is halved at $4, 6, 8$ epochs. We empirically set the balance parameters $\lambda_\mathrm{L1}, \lambda_\mathrm{ssim}, \lambda_\mathrm{smooth}, \lambda_\mathrm{consist}$ and $\lambda_\mathrm{reproj}$ in Eq.~\ref{equ:loss} to $1.0, 1.0, 1.0, 0.01, 1.0$ and $30$, respectively. All configurations and hyperparameters are shared for all experiments over three tasks unless specified.
% \vspace{-3mm}

\subsection{Depth Estimation}
% \vspace{-2mm}
We evaluate our depth estimation results on two standard benchmarks: ScanNet~\cite{dai2017scannet} and NYU-depth V2~\cite{Silberman_ECCV12}. 
% Both datasets contain indoor scenes with dense ground-truth depths. 
We use the synthesized (rendered) depth map as our prediction result and evaluated it by standard metrics introduced in~\cite{eigen2014depth}, including: absolute depth error (abs err), absolute relative depth error (abs rel), absolute log depth error (log$10$), squared relative error(sq rel), RMSE and inlier-ratio with threshold ($\sigma$). 

Given a testing frame, instead of using the monocular depth estimation results, we obtain the depth map via rendering with neural representation learnt from MonoNeRF. Comparing with the monocular depth predictions, the rendered depth maps are always more smooth. Before evaluation, we first align predictions with the ground truths for scale ambiguity issue, which is a common strategy for monocular depth estimation~\cite{tucker2020single, yin2021learning}. 

For the experiment on the ScanNet~\cite{dai2017scannet}, we train our framework with all training sequences and evaluate it on all testing sequences released in the official test split. We first compare our model with several fully supervised methods that trained with ground-truth depth supervision: MVDepthNet~\cite{wang2018mvdepthnet}, GPMVS~\cite{hou2019multi}, DPSNet~\cite{im2019dpsnet} and Atlas~\cite{murez2020atlas}.  We directly borrow the performance reported in their paper and list them in Table~\ref{sota:scannet}. Note that most of these methods are based on MVS with at least two images as input while our work only requires a single image as input. Without any depth ground truths, our approach still achieves a comparable result with some state-of-the-art. Meanwhile, compared to MonodepthV2~\cite{godard2019digging} which also only requires RGB supervision as ours, our method achieves much better performance.

\begin{table}[t]\setlength{\tabcolsep}{2pt}
\centering
% \scriptsize
\footnotesize
\begin{tabular} {l|cc|ccccc}
\toprule
Methods & Depth & Cam$_{ex}$ & Abs Rel$\downarrow$ & Sq Rel$\downarrow$  & RMSE $\downarrow$ & $\sigma1\uparrow$\\ \hline
\citet{wang2018mvdepthnet} & \cmark & \cmark & 0.098 &  0.061 & 0.293 & 89.6 \\
\citet{hou2019multi} & \cmark & \cmark & 0.130 &  0.339 & 0.472 & 90.6 \\
\citet{im2019dpsnet} & \cmark & \cmark & 0.087 & 0.035 & 0.232 & 92.5 \\
\citet{murez2020atlas} & \cmark & \cmark & 0.065 &  0.045 & 0.251 & 93.6 \\ \hline
% COLMAP~\cite{schonberger2016pixelwise} & -- & \xmark & \textbf{0.137} & \textbf{0.264} & 0.138 & 0.502 & \textbf{83.4} \\
\citet{godard2019digging} &  \xmark & \xmark & 
0.205 & 0.129 & 0.453 & 67.9 \\
MonoNeRF & \xmark & \xmark & \textbf{0.169} & \ \textbf{0.089} & \textbf{0.375} & \textbf{76.0} \\ 
\bottomrule
\end{tabular}
\vspace{-3mm}
\caption{\small{Comparison of depth estimation task on the Scannet~\cite{dai2017scannet} dataset. We measure the standard metrics on the whole test set.}}
\vspace{-4mm}
\label{sota:scannet}
\end{table}
\begin{table*}[t]
\setlength{\tabcolsep}{2pt}
% \small
\small
\centering
% \begin{tabular} {l|ccc|cccccc}
% \toprule
% Methods & Sup & Dataset & Cam$_{ex}$ & rel$\downarrow$ & log10$\downarrow$ & RMS$\downarrow$ & $\sigma1 \uparrow$ & $\sigma2 \uparrow$ & $\sigma3 \uparrow$ \\ \hline
% DIW~\cite{chen2016single} & Depth & DIW  & --  & 0.25 & 0.1 & 0.76 & 0.62 & 0.88 & 0.96 \\
% MegaDepth~\cite{li2018megadepth} & Depth & Mega & -- & 0.24 & 0.09 & 0.72 & 0.63 & 0.88 & 0.96\\
% 3DKenBurns~\cite{niklaus20193d} & Depth & Mega+NYU+3DKenBurn & -- & 0.08 & 0.03 & 0.30 & 0.94 & 0.99 & 1\\
% MiDaS~\cite{ranftl2020towards} & Depth & MiDaS 10 datasets & -- & 0.16 & 0.06 & 0.50 & 0.80 & 0.95 & 0.99\\
% \hline
% MPI~\cite{tucker2020single} & RGB$\dagger$ & RealEstate10K & \cmark & 0.15 & 0.06 & 0.49 & 0.81 & 0.96 & 0.99\\
% MINE~\cite{li2021mine} & RGB$\dagger$ & RealEstate10K & \cmark & 0.11 & 0.05 & 0.40 & 0.88 & 0.98 & 0.99 \\ \hline
% MonodepthV2~\cite{godard2019digging} & RGB & KITTI & \xmark & 0.25 & 0.10 & 0.74 & 0.62 & 0.87 & 0.95 \\
% MonodepthV2$^*$~\cite{godard2019digging} & RGB & RealEstate10K & \xmark & 0.31 & 0.12 & 0.82 & 0.51 & 0.83 & 0.91 \\
% Manydepth~\cite{watson2021temporal} & RGB & KITTI & \xmark & 0.25 & 0.10 & 0.76 & 0.61 & 0.87 & 0.95 \\
% Ours & RGB & RealEstate10K & \xmark & \textbf{0.17} & \textbf{0.07} & \textbf{0.57} & \textbf{0.73} & \textbf{0.94} & \textbf{0.98} \\
% \bottomrule
% \end{tabular}
\begin{tabular} {l|ccc|cccccc}
\toprule
Methods & Sup & Dataset & Cam$_{ex}$ & rel$\downarrow$ & log10$\downarrow$ & RMS$\downarrow$ & $\sigma1 \uparrow$ & $\sigma2 \uparrow$ \\ \hline
DIW~\cite{chen2016single} & Depth & DIW  & --  & 0.25 & 0.1 & 0.76 & 0.62 & 0.88  \\
MegaDepth~\cite{li2018megadepth} & Depth & Mega & -- & 0.24 & 0.09 & 0.72 & 0.63 & 0.88 \\
% 3DKenBurns~\cite{niklaus20193d} & Depth & Mega+NYU+3DKenBurn & -- & 0.08 & 0.03 & 0.30 & 0.94 & 0.99 \\
MiDaS~\cite{ranftl2020towards} & Depth & MiDaS 10 datasets & -- & 0.16 & 0.06 & 0.50 & 0.80 & 0.95 \\
\hline
MPI~\cite{tucker2020single} & RGB$\dagger$ & RealEstate10K & \cmark & 0.15 & 0.06 & 0.49 & 0.81 & 0.96 \\
MINE~\cite{li2021mine} & RGB$\dagger$ & RealEstate10K & \cmark & 0.11 & 0.05 & 0.40 & 0.88 & 0.98 \\ \hline
MonodepthV2~\cite{godard2019digging} & RGB & KITTI & \xmark & 0.25 & 0.10 & 0.74 & 0.62 & 0.87  \\
MonodepthV2$^*$~\cite{godard2019digging} & RGB & RealEstate10K & \xmark & 0.31 & 0.12 & 0.82 & 0.51 & 0.83 \\
Manydepth~\cite{watson2021temporal} & RGB & KITTI & \xmark & 0.25 & 0.10 & 0.76 & 0.61 & 0.87 \\
MovingIndoor~\cite{zhou2019moving} & RGB & NYU V2 & \xmark  & 0.21 & 0.09 & 0.71 & 0.67 & 0.90 \\ \hline
MonoNeRF & RGB & RealEstate10K & \xmark & \textbf{0.17} & \textbf{0.07} & \textbf{0.57} & \textbf{0.73} & \textbf{0.94} \\
\bottomrule
\end{tabular}
\vspace{-2mm}
\caption{\small{Comparison of depth estimation task on NYU Depth V2 dataset. We follow the standard metrics. ``Sup" denotes the supervision signal used during training. ``RGB$^\dagger$" means using both RGB image and sparse depth during training.
% , while our method only needs RGB images to achieve comparable performance. 
% NYU stands for whether the NYU Depth V2 data are used for training.
``MonodepthV2$^*$" is our reproduction of MonodepthV2 on RealEstate10K.}}
\vspace{-3mm}
\label{sota:nyu}
\end{table*}
Beyond ScanNet~\cite{dai2017scannet}, we also evaluate the depth estimation performance on NYU Depth V2~\cite{Silberman_ECCV12}. For a fair comparison, we train the model only with RealEstate10K~\cite{zhou2018stereo} training data as suggested in~\cite{tucker2020single, li2021mine} and report the results in Table~\ref{sota:nyu}. We split the existing method into three groups: (i) the depth supervision model: including Depth in the Wild~\cite{chen2016single}(DIW), MegaDepth~\cite{li2018megadepth}, 3DKenBurns~\cite{niklaus20193d} and MiDaS~\cite{ranftl2020towards}; (ii) the RGB supervision model with camera pose;
including MPI~\cite{tucker2020single} and MINE~\cite{li2021mine}; and (iii) the RGB supervision model without camera pose, including MonodepthV2~\cite{godard2019digging} and Manydepth~\cite{watson2021temporal}. 
Notably, compared to MiDas~\cite{ranftl2020towards} trained across $10$ different datasets with depth supervision, we achieve comparable performance. Although our method is slightly worse than MINE~\cite{li2021mine}, they utilize ground-truth camera poses for training while we do not. Compared with the approaches without neither depth supervision nor camera poses, our approach significantly outperforms them by a large margin.

% \input{figs/vis}

% \vspace{-2mm}
\begin{table}[t]
% \begin{table}
\setlength{\tabcolsep}{4pt}
\centering 
% \scriptsize
\footnotesize
\begin{tabular} {l|ccc}
\toprule
Methods & Mean$\downarrow$ & RMSE$\downarrow$ & Max err. $\downarrow$ \\ \hline
SSV~\cite{mustikovela2020self} & 0.142 & 0.175 & 0.365 \\
SfMLearner~\cite{zhou2017unsupervised} & 0.048 & 0.055 & 0.111 \\
P$^2$Net~\cite{yu2020p} & 0.059 & 0.068 & 0.148 \\
COLMAP~\cite{schonberger2016pixelwise} & 0.024 & 0.030 & 0.077 \\
VideoAE~\cite{lai2021video} & 0.017 & 0.019 & 0.041 \\ \hline
MonoNeRF & \textbf{0.009} & \textbf{0.011} & \textbf{0.022} \\ 
\bottomrule
\end{tabular}
\vspace{-2mm}
\caption{\small{Comparison of camera pose estimation task on RealEstate10K.}}
% We follow the pipeline in~\cite{lai2021video} and evaluate Absolute Trajectory Error (ATE).}}
% on 100 30-frame video clips and take an average across all clips.}
\vspace{-4mm}
\label{sota:cam}
% \end{table}
\end{table}
\subsection{Camera Pose Estimation}
% \vspace{-3mm}

We perform camera pose trajectory estimation and evaluate its performance on RealEstate10K~\cite{zhou2018stereo}. Following~\cite{lai2021video}, we use $1,000$ $30$-frames video clips from RealEstate10K testing data to construct the testing set. For each video clip, we take a pair of images as input and estimate the relative pose between them and repeat this step sequentially through the whole video to obtain the full trajectory. Since the model only estimates the relative pose in the world coordinate defined in our model, we adopt a post-processing step for alignment between the predicted camera trajectory and the SfM trajectory provided by RealEstate10K~\cite{zhou2018stereo} via the Umeyama algorithm~\cite{umeyama1991least}. We evaluate the Absolute Trajectory Error (ATE) over testing videos and compare it with the state-of-the-art methods in Table~\ref{sota:cam}.

\begin{figure*}[t]
    \centering
    % \vspace{-2mm}
	\includegraphics[width=0.9\textwidth]{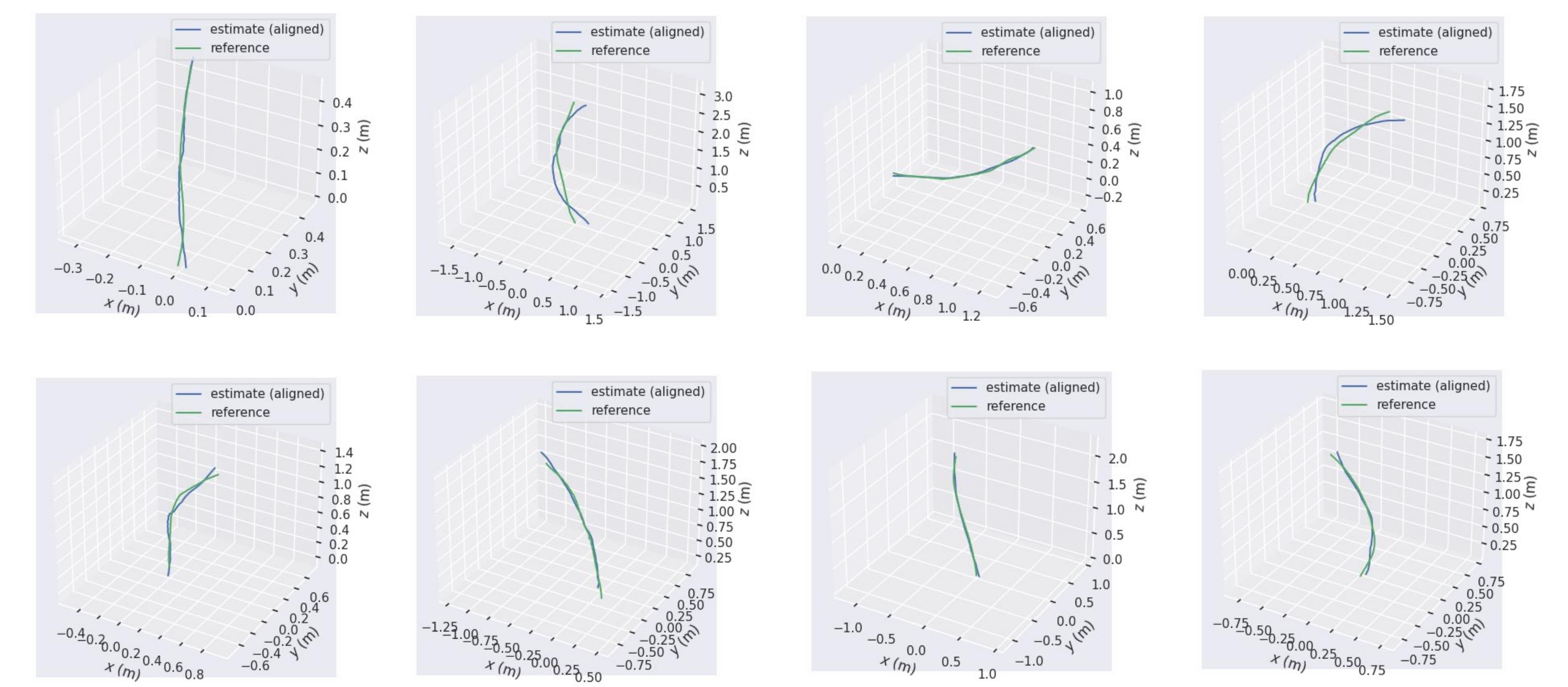}
	\vspace{-2mm}
	\caption{\small{Visualization of estimated camera trajectory on RealEstate10K~\cite{zhou2018stereo}. The green trajectory indicates the ground-truth camera poses while the blue one indicates the estimated poses.}}
	\label{fig:cam}
    \vspace{-2mm}
\end{figure*}

SfMLearner~\cite{zhou2017unsupervised} and P$^2$Net~\cite{yu2020p} are two works related to ours, which borrow similar ideas from traditional SfM and optimize the camera trajectory and depth map jointly. 
% We apply the same alignment step between the predictions and ground truths. 
Our approach outperforms them by a large margin. For instance, the RMSE is reduced from $0.055$ to $0.011$ which is about a $80\%$ improvement. In addition, our approach is superior compared to the COLMAP~\cite{schonberger2016pixelwise} based on the SfM pipeline. Especially for the videos with slow and little camera movement, COLMAP~\cite{schonberger2016pixelwise} can hardly work well and always requires plenty of frames to process leading to a much longer inference time. 
% Moreover, we compare our approach with the self-supervised viewpoint estimation approach SSV~\cite{mustikovela2020self}. Our approach significantly reduces the mean error by $10$ times.
Finally, a similar improvement can be also found when comparing to VideoAE~\cite{lai2021video}, which is a recent work on the disentanglement of camera motion and 3D structure. The qualitative result is shown in Fig.~\ref{fig:cam}

\begin{table}[t]
\setlength{\tabcolsep}{2pt}
\centering
% \scriptsize
\footnotesize
% \small
% \vspace{-3mm}
\begin{tabular} {l|c|ccc}
\toprule
Methods & Cam$_{ex}$ & PSNR$\uparrow$ & SSIM$\uparrow$ & Perc Sim$\downarrow$ \\ 
\hline
\citet{dosovitskiy2015learning} & \cmark & 11.35 & 0.33 & 3.95 \\
GQN~\cite{eslami2018neural} & \cmark & 16.94 & 0.56 & 3.33 \\
\citet{zhou2016view} & \cmark & 17.05 & 0.56 & 2.19 \\
GRNN~\cite{tung2019learning} & \cmark & 19.13 & 0.63 & 2.83 \\
% 3DPaper~\cite{wiles2020synsin} & \cmark & 21.88 & 0.66 & 1.52 \\
SynSin(w/ voxel)  & \cmark & 21.88 & 0.71 & 1.30 \\
SynSin~\cite{wiles2020synsin}  & \cmark & 22.31 & 0.74 & \textbf{1.18} \\
StereoMag$\dagger$~\cite{zhou2018stereo} & \cmark & \textbf{25.34} & \textbf{0.82} & 1.19 \\ 
\hline
SSV~\cite{mustikovela2020self} & \xmark & 7.95 & 0.19 & 4.12 \\
SfMLearner~\cite{zhou2017unsupervised} & \xmark & 15.82 & 0.46 & 2.39 \\
MonoDepth2~\cite{godard2019digging} & \xmark & 17.15 & 0.55 & 2.08 \\
P$^2$Net~\cite{yu2020p} & \xmark & 17.77 & 0.56 & 1.96 \\
VideoAE~\cite{lai2021video} & \xmark & 23.21 & 0.73 & 1.54 \\
MonoNeRF & \xmark & \textbf{25.00} & \textbf{0.83} & \textbf{0.99} \\
\hline \hline
\multicolumn{5}{l}{\textit{results on the test split proposed in~\cite{li2021mine}}} \\
% \hline
MPI$\ddagger$~\cite{tucker2020single}  & \cmark & 27.05 & 0.87 & 0.097$^*$ \\
MINE$\ddagger$~\cite{li2021mine}  & \cmark & 28.39 & 0.90 & 0.090$^*$ \\ \hline
MonoNeRF & \xmark & 26.68 & 0.86 & 0.143$^*$ \\ 
\bottomrule
\end{tabular}
\vspace{-2mm}
\caption{\small{Comparison of novel view synthesis task on RealEstate10K. We follow the standard metrics of PSNR, SSIM, and Perc Sim~\cite{wiles2020synsin}. The number $\mathrm{xx}^*$ represents the LPIPS metric using the implementation of~\cite{zhang2018perceptual}.$\dagger$StereoMag makes use of 2 images as input. $\ddagger$MPI and $\ddagger$MINE use sparse point clouds as the additional supervision signal during training.}}
\vspace{-6mm}
\label{sota:nvs}
\end{table}

\begin{table}[t]
  \centering
  % \small
  \vspace{-2mm}
  \footnotesize
  \setlength{\tabcolsep}{4pt}
  \begin{tabular}{cccc}
    \toprule
    calib.  & PSNR$\uparrow$ & SSIM$\uparrow$ & LPIPS$\downarrow$   \\
    \midrule
    \xmark  & 21.46 & 0.677 & 0.289 \\
    \cmark  & 26.68 & 0.863 & 0.143 \\
    % \bottomrule
    \bottomrule
  \end{tabular}
  \vspace{-3mm}
    \caption{\small{Novel view synthesis on RealEstate10K w./w.o. auto scale calibration (Sec.~\ref{sec:scalecalibration}).}}
      \label{tab:calib}
      \vspace{-2mm}
\end{table}
\begin{table}[t]
  \centering
  % \small
  \footnotesize
  \setlength{\tabcolsep}{4pt}
  \begin{tabular}{lccc}
    \toprule
    \#D  & PSNR$\uparrow$ & SSIM$\uparrow$ & LPIPS$\downarrow$   \\
    \midrule
    64  & 26.68 & 0.863 & 0.143 \\
    32  & 26.56 & 0.861 & 0.141 \\
    16  & 26.65 & 0.861 & 0.144 \\
    \bottomrule
  \end{tabular}
  \vspace{-3mm}
  \caption{\small{Novel view synthesis on RealEstate10K with the different number of planes.}}
  \vspace{-3mm}
\label{tab:plane}
\end{table}
\begin{table}[t]
  \centering
  % \scriptsize
  \footnotesize
  \setlength{\tabcolsep}{4pt}
  \begin{tabular}{cccc}
    \toprule
    ratio  & PSNR$\uparrow$ & SSIM$\uparrow$ & LPIPS$\downarrow$   \\
    \midrule
    1.0  & 26.68 & 0.863 & 0.143 \\
    0.8  & 26.61 & 0.860 & 0.145 \\
    0.6 & 26.31 & 0.857 & 0.147 \\
    0.4 & 26.21 & 0.851 & 0.151 \\
    0.2 & 25.52 & 0.834 & 0.161 \\
    % \bottomrule
    \bottomrule
  \end{tabular}
  \vspace{-3mm}
   \caption{\small{Novel view synthesis on RealEstate10K with different ratios of training data.}}
     \label{tab:data}
     \vspace{-3mm}
\end{table}
% \begin{table*}[t]
% \setlength{\tabcolsep}{3pt}
% \centering
% % \footnotesize
% \small
% \begin{tabular} {l|ccc|ccc}
% \toprule
% \multirow{2}{*}{Methods}& \multicolumn{3}{c|}{\textit{novel view synthesis}} & \multicolumn{3}{c}{\textit{depth estimation}} \\ 
%  & PSNR$\uparrow$ & SSIM $\uparrow$ & Perc Sim$\downarrow$ & Abs Rel$\downarrow$ & Sq Rel$\downarrow$ & RMSE$\downarrow$ \\ \hline
% Appearance Flow & 14.8 & 0.48 & 3.13 & -- & -- & -- \\
% SynSin & 15.7 & 0.47 & 2.76 & 0.91 & 1.81 & 2.08 \\
% MINE & \textbf{19.3} & \textbf{0.71} & \textbf{1.69} & 0.19 & 0.18 & \textbf{0.34} \\ \hline
% Ours & 18.0 & 0.61 & 2.11 & \textbf{0.17} & \textbf{0.09} & 0.39\\ 
% \bottomrule
% % Methods & Mean$\downarrow$ & RMSE$\downarrow$ & Max err. $\downarrow$ \\ \hline
% % SSV~\cite{mustikovela2020self} & 0.142 & 0.175 & 0.365 \\
% % SfMLearner~\cite{zhou2017unsupervised} & 0.048 & 0.055 & 0.1105 \\
% % P$^2$Net~\cite{yu2020p} & 0.059 & 0.068 & 0.1475 \\
% % COLMAP~\cite{schonberger2016pixelwise} & 0.024 & 0.030 & 0.0765 \\
% % VideoAE~\cite{lai2021video} & 0.017 & 0.019 & 0.0410 \\
% % Ours & \textbf{0.009} & \textbf{0.011} & \textbf{0.0223} \\ \hlineB{2.5}
% \end{tabular}
% \vspace{-3mm}
% \caption{\footnotesize{Generalization ability of novel view synthesis task and depth estimation task. We pretrain our model on the RealEstate10K~\cite{zhou2018stereo} and evaluate on the 100 30-frames clips of ScanNet~\cite{dai2017scannet}.}}
% \vspace{-6mm}
% \label{tab:generalize}
% \end{table*}

\begin{table}[t]
\setlength{\tabcolsep}{3pt}
\centering
% \scriptsize
\footnotesize
% \small
\begin{tabular} {l|ccc}
\toprule
\multirow{2}{*}{Methods}& \multicolumn{3}{c}{\textit{novel view synthesis}} \\
 & PSNR$\uparrow$ & SSIM $\uparrow$ & Perc Sim$\downarrow$  \\ \hline
Appearance Flow~\cite{zhou2016view} & 14.8 & 0.48 & 3.13  \\
Synsin~\cite{wiles2020synsin} & 15.7 & 0.47 & 2.76  \\
MINE~\cite{li2021mine} & \textbf{19.3} & \textbf{0.71} & \textbf{1.69}  \\ \hline
MonoNeRF & 18.0 & 0.61 & 2.11 \\ 
\bottomrule
\multirow{2}{*}{Methods} & \multicolumn{3}{c}{\textit{depth estimation}} \\ 
& Abs Rel$\downarrow$ & Sq Rel$\downarrow$ & RMSE$\downarrow$ \\ \hline
Appearance Flow~\cite{zhou2016view} &  -- & -- & -- \\
Synsin~\cite{wiles2020synsin} &  0.91 & 1.81 & 2.08 \\
MINE~\cite{li2021mine} & 0.19 & 0.18 & \textbf{0.34} \\ \hline
MonoNeRF & \textbf{0.17} & \textbf{0.09} & 0.39\\ \bottomrule
% Methods & Mean$\downarrow$ & RMSE$\downarrow$ & Max err. $\downarrow$ \\ \hline
% SSV~\cite{mustikovela2020self} & 0.142 & 0.175 & 0.365 \\
% SfMLearner~\cite{zhou2017unsupervised} & 0.048 & 0.055 & 0.1105 \\
% P$^2$Net~\cite{yu2020p} & 0.059 & 0.068 & 0.1475 \\
% COLMAP~\cite{schonberger2016pixelwise} & 0.024 & 0.030 & 0.0765 \\
% VideoAE~\cite{lai2021video} & 0.017 & 0.019 & 0.0410 \\
% Ours & \textbf{0.009} & \textbf{0.011} & \textbf{0.0223} \\ \hlineB{2.5}
\end{tabular}
\vspace{-2mm}
\caption{\small{Generalization ability of novel view synthesis task and depth estimation task. We pretrain our model on the RealEstate10K and evaluate on the 100 30-frames clips of ScanNet.}}
\vspace{-5mm}
\label{tab:generalize}
\end{table}

\subsection{Novel View Synthesis}

Our approach generates novel view images by rendering the Multiplane NeRF representation into target views. Following the setting of~\cite{wiles2020synsin, lai2021video}, we evaluate the novel view synthesis on RealEstate10K~\cite{zhou2018stereo}, which is a large-scale walkthrough video dataset with both indoor and outdoor scenes. During testing, we follow two test splits provided by SynSin~\cite{wiles2020synsin} and MINE~\cite{li2021mine}. For evaluation, we randomly sample $5$ source frames from each testing sequence and sample target frames that are $5$ frames apart from the source frames. We measure the similarity scores by PSNR, SSIM~\cite{wang2004image}, and perceptual similarity with VGG~\cite{simonyan2014very} features. Note that there are two different implementations to calculate the perceptual similarity used in SynSin and MINE, the latter one is also known as LPIPS~\cite{zhang2018perceptual}. Table~\ref{sota:nvs} summarizes the novel view synthesis performance over different methods. Compared to single-image view synthesis algorithms, for instance, Synsin~\cite{wiles2020synsin} , our method can achieve comparable or better performance, even though our method does not require camera pose ground truths while other methods do. 
For instance, our method is better than Synsin~\cite{wiles2020synsin} over all three metrics. Some qualitative results are shown in Fig.~\ref{fig:vis} and more can be found in the supplementary. 

Compared with MPI~\cite{tucker2020single} and MINE~\cite{li2021mine} where similar 3D representations are adopted, our approach is slightly worse on PSNR and SSIM. We believe this inferior performance is reasonable since they rely on the ground-truth camera pose and the sparse points obtained by COLMAP~\cite{schonberger2016pixelwise} during training and testing. On the other hand, our approach easily outperforms all existing methods of training without the camera pose. Some qualitative results are shown in Fig.~\ref{fig:vis} and more can be found in the supplementary material.

\begin{figure*}[t]
    \centering
	\includegraphics[width=0.85\textwidth]{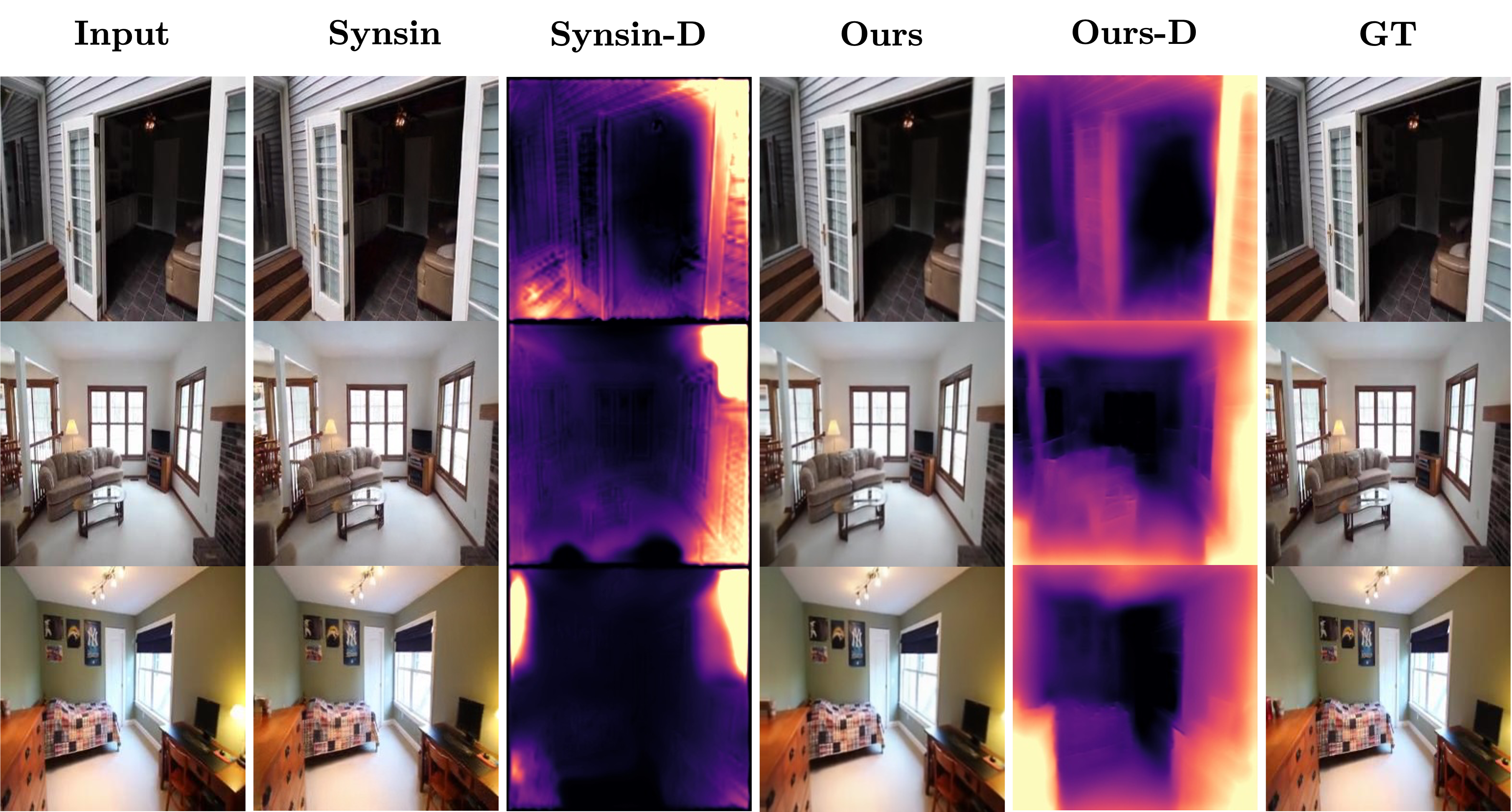}
	\vspace{-2mm}
	\caption{\small{Visualization of depth and novel view images on RealEstate10K. We compare our method with Synsin. Despite they share similar quality of generated images, our depth output is much more accurate.}}
	\label{fig:vis}
	\vspace{-3mm}
\end{figure*}
\subsection{Ablation Study}

We find the performance of three tasks are aligned in our experiments, thus we report the ablation mainly based on the novel view synthesis task here. 
% \input{tables/ablation}

% Please see supplementary material for more experiments.
% \input{tables/calib}
\textbf{Auto scale calibration.} We show the effectiveness of auto-calibration (Sec.~\ref{sec:scalecalibration}) by conducting an experiment w./w.o the calibration step. As shown in Table~\ref{tab:calib}, the novel view synthesis performance drops dramatically without the auto scale calibration, \ie, more than $5\%$ on PSNR, which indicates this calibration step is beneficial to scale-invariant synthesis.

\textbf{Number of planes.} We compare our default model with different numbers of planes used in Multiplane NeRF as listed in Table~\ref{tab:plane}. We found that our approach is not so sensitive to the number of planes, but in general, our default setting achieves the best performance.

\textbf{Amount of training data.} We analyze the effect of using different fractions of training data. We uniformly sample every 20\% fraction of RealEstate10K~\cite{zhou2018stereo} training data and evaluate the performance on the same test set. As reported in Table~\ref{tab:data}, with more training data, the quality of generated images is getting better.

\textbf{Generalization ability.} To show the generalization ability of our model, we utilize the model pretrained on RealEstate10K and evaluate the performance of novel view synthesis and depth estimation on ScanNet. As illustrated in Table~\ref{tab:generalize}, our model can achieve on par or even better results on both two tasks.

\section{Conclusion}
\label{sec:conclusion}
We present an autoencoder architecture that disentangles video into camera motion and depth map via the camera encoder and the depth encoder. And the Multiplane NeRF is utilized as the decoder to represent the 3D scene. We further introduce an auto-scale calibration strategy to learn the disentanglement representation even with the camera pose. With the powerful 3D representation, we show our model enables camera pose estimation, depth estimation, and novel view synthesis. Our model achieves on-par or even better results on three tasks compared to approaches with the ground-truth camera or depth during training.

{\footnotesize \noindent \textbf{Acknowledgements.} This project was supported, in part, by NSF CCF-2112665 (TILOS), NSF CAREER Award IIS-2240014, NSF 1730158 CI-New: Cognitive Hardware and Software Ecosystem Community Infrastructure (CHASE-CI), NSF ACI-1541349 CC*DNI Pacific Research Platform, Amazon Research Award, Sony Research Award, Adobe Data Science Research Award, and gifts from Qualcomm and Meta.}


\bibliography{example_paper}
\bibliographystyle{icml2023}

\end{document}